\documentclass[letterpaper]{article} 
\usepackage{aaai24}  
\usepackage{times}  
\usepackage{helvet}  
\usepackage{courier}  
\usepackage[hyphens]{url}  
\usepackage{graphicx} 
\usepackage{natbib}  
\usepackage{caption} 
\usepackage{algorithm}
\usepackage{algorithmic}
\usepackage{amsfonts}
\usepackage{amsmath}
\usepackage{lineno}
\usepackage{newfloat}
\usepackage{listings}
\DeclareCaptionStyle{ruled}{labelfont=normalfont,labelsep=colon,strut=off} 
\floatstyle{ruled}
\newfloat{listing}{tb}{lst}{}
\floatname{listing}{Listing}
\usepackage{tikz,graphics,color,float,epsf,caption,subcaption,enumitem}
\usepackage{xcolor}

\title{Multi-Prompts Learning with Cross-Modal Alignment for\\
Attribute-based Person Re-Identification}

\author{
    Yajing Zhai\textsuperscript{\rm 1,2}\thanks{These authors contributed equally. This work was done when Yajing Zhai was an intern at Eastern Institute of Technology, Ningbo, China.},
    Yawen Zeng\textsuperscript{\rm 1}\footnotemark[1],
    Zhiyong Huang\textsuperscript{\rm 3},
    Zheng Qin\textsuperscript{\rm 1}\thanks{Corresponding authors.},
    Xin Jin\textsuperscript{\rm 2}\footnotemark[2],
    Da Cao\textsuperscript{\rm 1}
}

\affiliations{
    \textsuperscript{\rm 1}College of Computer Science and Electronic
Engineering, Hunan University, Changsha, China \\

    \textsuperscript{\rm 2}Ningbo Institute of Digital Twin, Eastern Institute of Technology, Ningbo, China \\

    \textsuperscript{\rm 3}National University of Singapore, NUS Research Institute in Chongqing\\ 
        
    {\texttt\small \{yajingzhai9,yawenzeng11\}@gmail.com, huangzy@comp.nus.edu.sg,\\
    zqin@hnu.edu.cn, jinxin@eitech.edu.cn, caoda0721@gmail.com}
}

\usepackage{bibentry}

\begin{document}

\maketitle

\begin{abstract}
The fine-grained attribute descriptions can significantly supplement the valuable semantic information for person image, which is vital to the success of person re-identification (ReID) task. However, current ReID algorithms typically failed to effectively leverage the rich contextual information available, primarily due to their reliance on simplistic and coarse utilization of image attributes. Recent advances in artificial intelligence generated content have made it possible to automatically generate plentiful fine-grained attribute descriptions and make full use of them. Thereby, this paper explores the potential of using the generated multiple person attributes as prompts in ReID tasks with off-the-shelf (large) models for more accurate retrieval results. To this end, we present a new framework called Multi-Prompts ReID (MP-ReID), based on prompt learning and language models, to fully dip fine attributes to assist ReID task. Specifically, MP-ReID first learns to hallucinate diverse, informative, and promptable sentences for describing the query images. This procedure includes (i) explicit prompts of which attributes a person has and furthermore (ii) implicit learnable prompts for adjusting/conditioning the criteria used towards this person identity matching. Explicit prompts are obtained by ensembling generation models, such as ChatGPT and VQA models. Moreover, an alignment module is designed to fuse multi-prompts (i.e., explicit and implicit ones) progressively and mitigate the cross-modal gap. Extensive experiments on the existing attribute-involved ReID datasets, namely, Market1501 and DukeMTMC-reID, demonstrate the effectiveness and rationality of the proposed MP-ReID solution.
\end{abstract}

\section{Introduction}

Person re-identification (ReID) is a challenging task due to the dramatic visual appearance changes from pose, viewpoints, illumination, occlusion, low resolution, background clutter, etc. \cite{jin2020uncertainty,ye2021deep,zhang2021person}. Fine-grained person attributes are robust to these variations and are often exploited as efficient supplements with local descriptions that aid in the learning of more discriminative feature representations \cite{jia2022learning,wang2022pedestrian}. In particular, the common attributes include clothing color, shoes, hairstyle, gender, age, and other specific characteristics. They serve as additional information that complements and aligns images, reducing the impact of the above factors, thereby improving the overall performance of ReID~\cite{yu2022multi}. 

\begin{figure}[t!]
  \centering
  \begin{subfigure}[t]{1\columnwidth}
		\centering
		\includegraphics[width=0.99\linewidth]{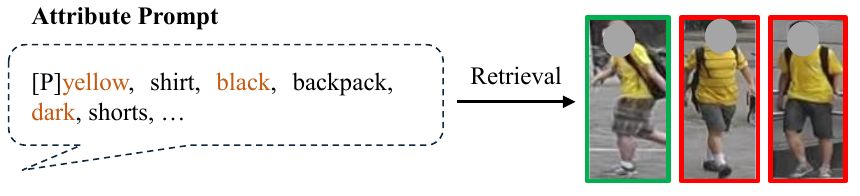}
        \vspace{-0.1cm}
		\caption{Retrieval results with limited coarse-grained attributes} 
        \vspace{0.1cm}
    \label{subfig:example_a}
  \end{subfigure}
  \begin{subfigure}[t]{1\columnwidth}
		\centering
		\includegraphics[width=0.99\linewidth]{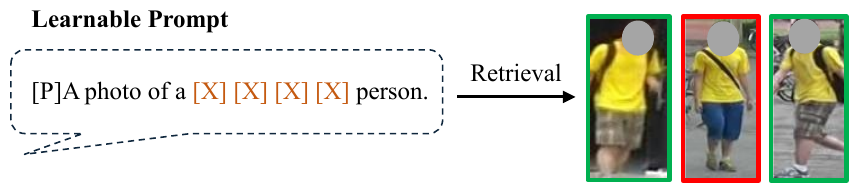}
        \vspace{-0.1cm}
        \caption{Retrieval results with only implicit attribute prompts}
        \vspace{0.1cm}
    \label{subfig:example_b}
  \end{subfigure}
  \begin{subfigure}[t]{1\columnwidth}
		\centering
		\includegraphics[width=0.99\linewidth]{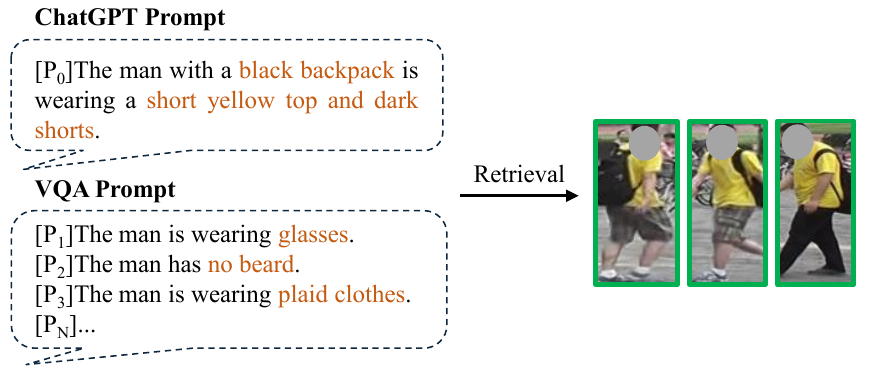}
		\vspace{-0.1cm}
        \caption{Retrieval results with multiple fine-grained attribute prompts}
    \label{subfig:example_c}
  \end{subfigure}
   \vspace{-0.3cm}
   \caption{Comparison of various usages of attributes for person ReID, where red boxes represent negative images, green boxes indicate positive results and key attribute words have been marked in red. We can see that using multiple fine human attributes as prompts in ReID brought advancements.}
   \vspace{-0.4cm}
	\label{fig:example}
\end{figure}

Recently, some researchers \cite{jia2021spatial,niu2022cross,specker2023upar, zheng2022progressive} begin to investigate the importance of attributes w.r.t ReID task, which demonstrate that attributes are indeed an effective piece of information that could enhance the retrieval performance in ReID. 
However, current attribute-based ReID algorithms fail to leverage the full potential of the abundant contextual information available. That's mainly because they rely on simplistic and naive utilization of coarse-grained attributes, as well as the complexity of accurately capturing and descriptions with the limitation of AI technology in the past. As shown in Figure~\ref{subfig:example_a}, certain coarse, separate, and ambiguous attributes, such as ``yellow", ``shorts", and ``shirt", are directly used to retrieve pedestrians, which are less effective compared to clear and complete abundant contextual descriptions as presented in Figure~\ref{subfig:example_c}. Thus, it is essential but has not been well investigated to efficiently take full advantage of fine-grained attribute information for improving ReID accuracy.

With the fast-development of large models \cite{fu2022large,jin2022meta,li2023blip2}, ReID methods gradually become more practical for real-world scenarios and gain superior performance. Besides, prompt learning~\cite{wu2022fast,zhou2022learning}, as a paradigm of strategies that leverages pre-trained models by incorporating additional textual description information, has achieved improved performance in many complex AI tasks~\cite{zeng2022point,luddecke2022image,liu2022dpt}. Building upon this inspiration, we investigate the feasibility of utilizing prompts to provide fine-grained attribute information for the ReID task.

Intuitively, there are two strategies for applying attributes as prompts, explicit attribute prompts and implicit attribute prompts, as shown in Figure~\ref{fig:example}. (i) Explicit attribute prompts refer to an attribute-based sentence prompt generation method, where the production process utilizes some attribute words, among which ChatGPT and visual question answer (VQA) models \cite{yu2019mcan, Wang2022vqa} with stronger interactivity and feedback mechanism are often used. (ii) While implicit attribute prompts use a learnable textual prompt generation method, where the process does not have specific attribute information, as depicted in Figure~\ref{subfig:example_b}. We can see that, the better retrieval result is obtained via the implicit attribute prompt method, but it is still not accurate enough. In contrast, as shown in Figure~\ref{subfig:example_c}, the ReID scheme that learning from multiple attribute prompts significantly improves the retrieval performance with more fine-grained information. From this we can infer that, the utilization of fine attribute information could enable the ReID model to learn more auxiliary features and relationships, thereby improving ultimate accuracy.

However, prominent challenges still remain that need to be further addressed. Firstly, the lack of such a required prompt-related ReID dataset in the large-scale practical ReID task has led to few studies have been exploited. The second challenge is that there is a gap between the attribute-based text prompt and the image, making it essential to address the alignment of these two modalities. As a result, despite utilizing rich prompts for improved ReID performance is a promising approach that can lead to efficient and comprehensive results, it remains an under-explored area with the potential for further optimization.

In this paper, we make the first attempt to employ the large-scale multi-prompts information in the attribute-based ReID task and propose a novel \textbf{M}ulti-\textbf{P}rompts Learning framework, dubbed as \textbf{MP-ReID}, to support this challenging task. MP-ReID aims to retrieve one person based on a variety of fine-grained attribute information as a complement for image information to improve the retrieval performance with ChatGPT, VQA and CLIP \cite{radford2021learning}. As mentioned above, the multi-prompts include explicit attribute prompts and implicit attribute prompts. 1) \textbf{Explicit attribute prompts} --- a prompt sentence generation paradigm, which is ensembling generation models based on attribute words. 2) The other is \textbf{implicit attribute prompts} --- a learnable prompt paradigm without intuitive attributes, which models a prompt’s context words with learnable vectors, that could be initialized with either random values or pre-trained word embeddings. Furthermore, image information with the promptable semantic feature is optimized under a cross-modal space to mitigate the cross-modal semantic gap. After that, the learned prompts are regarded as a booster to apply to the person retrieval. By conducting experiments on two well-known datasets, we validate that MP-ReID is superior to various existing methods. The main contributions of this work are summarized as follows:
\begin{itemize}[leftmargin=*]
\item{To the best of our knowledge, this is the first attempt that introduces the concept of multi-prompts learning strategies to generate diverse, informative, and promptable sentences for ReID improvements.} 
\item{We introduce two prompts generation paradigms: explicit attribute prompt and implicit attribute prompt, for applying fine-grained attributes to fully use the comprehensive semantics and enhance the retrieval performance.}
\item{We contribute a Multi-Prompts ReID framework, dubbed MP-ReID to mitigate the cross-modal semantic gap for this attribute-based ReID task. Meanwhile, we collect a prompt-related ReID dataset containing multiple attribute prompts about the same person, and we have released the dataset to facilitate the research community\footnote{https://github.com/zyj20/MPReID.}.}
\end{itemize}

\begin{figure*}[t]
    \centering
    \includegraphics[width=\linewidth]{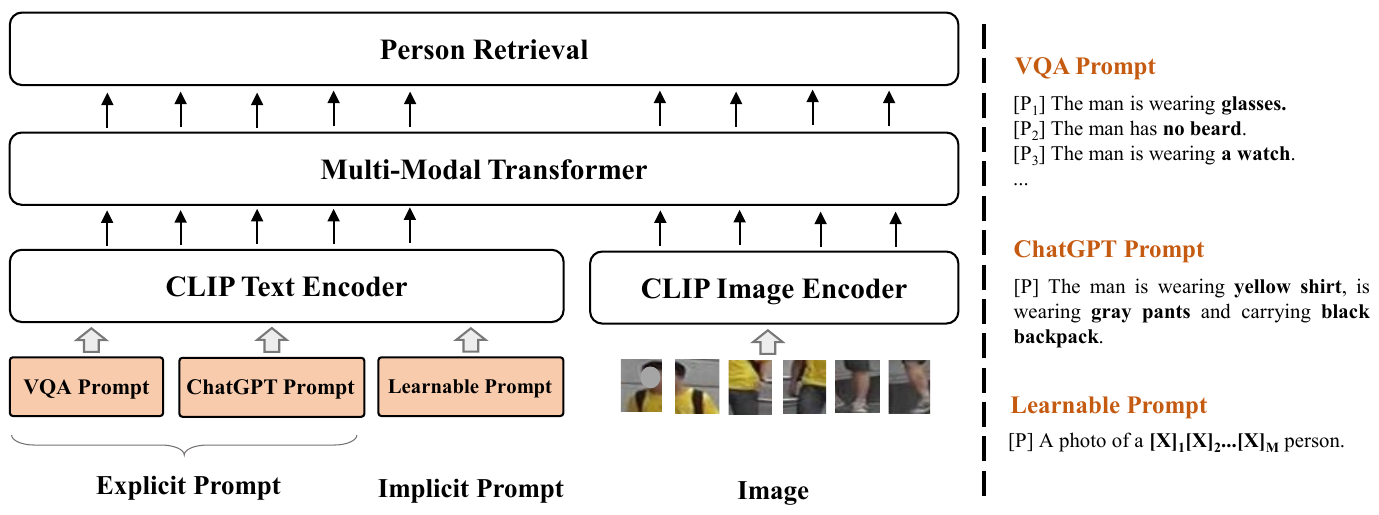}
    \caption{The graphical representation of MP-ReID for ReID. Under the prompt learning paradigm, the multi-prompts generated by ChatGPT and VQA are regarded as the textual input to the multi-modal Transformer, which can enhance the retrieval of the matching person images. It is built upon three components: 1) Multi-Prompts Generation Learning; 2) Cross-Modal Alignment; 3) Person Retrieval.}
    \label{frameangel}
\end{figure*}

\section{Related Work}
\subsection{Prompt Learning}

In recent years, the use of prompt learning, which concerns providing suggestive information, has become a popular technique for incorporating knowledge in natural language processing problems \cite{petroni2019language, song2022v2p,jin2023domain}. It involves adding language-specific instructions to the input text, enabling the pre-trained model to comprehend the downstream task and enhance the performance. ChatGPT\footnote{\url{https://openai.com/blog/chatgpt}} and GPT-4\footnote{\url{https://openai.com/product/gpt-4}} offer tremendous opportunities to improve open-source large language models using instruction-tuning \cite{peng2023instruction} and transfer to downstream tasks with powerful generalization \cite{zhang2023one}. Moreover, there has been a recent trend towards utilizing prompt learning for improving the quality of visual representations in vision-language models \cite{ju2022prompting, rao2022denseclip}. 

\subsection{Attributed-based Person ReID}

Recently, some deep learning methods are proposed to exploit the discriminative attributes. In particular, Li et al. \cite{li2019attribute}  manually labeled pedestrian attributes for the Market1501 dataset and the DukeMTMC-reID dataset\footnote{\url{https://vana77.github.io}}. Besides, the authors proposed a novel attribute-based person recognition framework with an attribute re-weighting module. This aims to learn discriminative embedding and correct prediction. Zhang et al. \cite{zhang2020person} leveraged the feature aggregation strategy to make use of attribute information. Jeong et al. \cite{jeong2021asmr} presented a new loss for learning cross-modal embeddings in the context of attribute-based person search and regarded attribute dataset as a category of people sharing the same traits. Li et al. \cite{li2022clip} fully exploited the cross-modal description ability through a set of learnable text tokens for each person ID and gave them to the text encoder to form ambiguous descriptions with a two-stage strategy, facilitating a better visual representation.

Inspired by the above work, we optimize the descriptive and visual features under the multi-prompts generation paradigm for ReID task, which contains explicit prompts and implicit prompts. In this way, textual prompts and visual features are learned from each other, achieving a win-win effect.

\section{Methodology} \label{methodology}
This section provides a detailed explanation of our solution, with Figure~\ref{frameangel} illustrating the overall framework of our MP-ReID. Generally speaking, our proposed framework comprises three components: multi-prompts generation, cross-model alignment, and person retrieval. 
1). The approach of multi-prompts generation learning leverages ChatGPT, VQA, and learnable methods to generate three different prompts, which is given in Figure~\ref{frameprompt}. These prompts are then fused together using a cross-attention mechanism; 2). Cross-modal alignment module aligns prompts-images pairs by feeding them into a multi-modal Transformer to learn the context; and 3). Person retrieval involves creating a feature representation in the prompt-visual space for identifying individuals.

\subsection{Multi-Prompts Generation Learning}\label{mpgl}
Given an attribute-based ReID dataset, images are defined as $\mathbb{M}=\{m_1, m_2, ..., m_n\}$, the corresponding attributes are denoted as $\mathbb{A}=\{a_1, a_2, ..., a_n\}$, respectively. The MP-ReID first generates prompts $\mathbb{P}=\{p_1, p_2, ..., p_n\}$, which contains $P_{i}^{e}$ ensembling explicit prompts and $P_{i}^{l}$ learnable implicit prompts. For visual representation and prompt representation, we adopt the image encoder and the text encoder from CLIP as the backbone for feature extractor respectively. They are all implemented as Transformer architecture~\cite{he2021dense,zhu2023learning}. And the ViT-B/16 network architecture \cite{dosovitskiy2020vit} is utilized for the images, which contains $12$ transformer layers. With respect to prompts embedding, we convert each word into a unique numeric ID using byte pair encoding with a $49,512$ vocab size \cite{sennrich2016neural}. To enable parallel computation, we set the context length of each text sequence to $77$, including the start [SOS] and end [EOS] tokens. Within a batch of images, we denote the index of each image as $i \in \{ {1...N} \}$. We calculate the similarity between the [CLS] token embedding of the image feature $m_{i}$ and the corresponding [EOS] token embedding of the text feature $p_{i}$. And in this module, we obtain the image feature 
\begin{equation}\label{eqn1}
f_{i}^{m} = F_m(m_i)
\end{equation}
Accordingly, for textual representation, we obtain the prompt feature $f_{i}^{p}$, which is formally formulated as,
\begin{equation}\label{eqn2}
{f}_{i}^{p}\leftarrow
\left\{
\begin{array}{l}
{f}_{i}^{e} = F_p(p_{i}^{e})\\
{f}_{i}^{l} = F_p(p_{i}^{l})
\end{array}
\right.
\end{equation}where $F_m (\cdot)$ and $F_p (\cdot)$ are visual and textual projection function. Besides, $f_{i}^{m}$, ${f}_{i}^{e}$, ${f}_{i}^{l}$ are extracted image features, explicit prompt features and implicit prompt features, respectively. $P_{i}^{e}$ ensembling explicit prompts comprise $P_{i}^{c}$ ChatGPT generation prompt and $P_{i}^{v}$ VQA generation prompt.

\begin{figure}[t]
    \centering
    \label{frameprompt}
    \includegraphics[width=\linewidth,height=6.5cm]{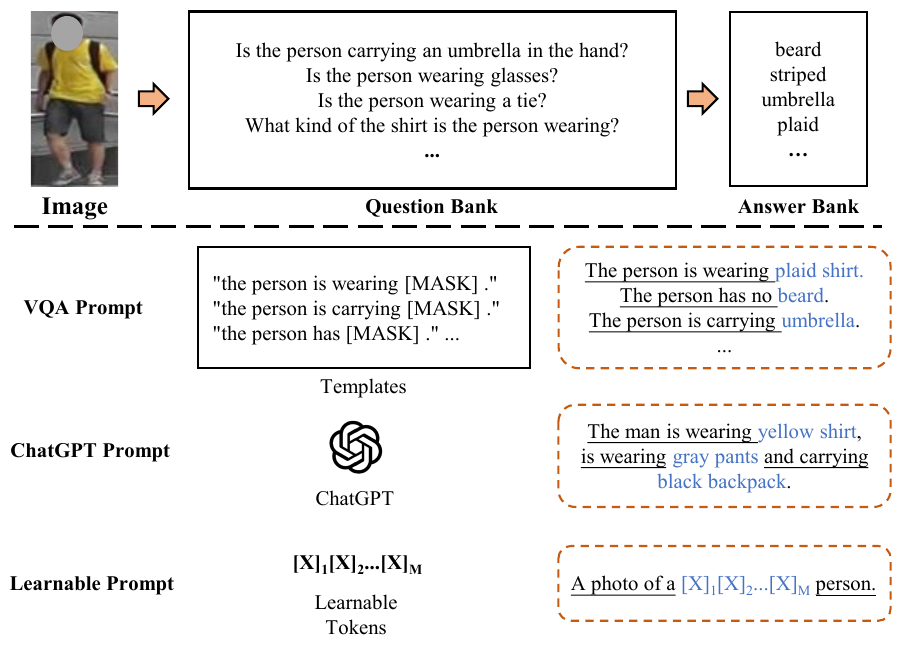}
    \caption{The process of multi-prompts generation learning in the proposed MP-ReID framework. The top is the question bank and answer bank of VQA model; and the bottom is the concrete multi-prompts generation from the VQA prompt, ChatGPT prompt and learnable prompt, respectively.}
    \label{frameprompt}
\end{figure}

\textbf{1) Explicit attribute prompts.} Specifically, to generate explicit attribute prompts, we adopt a prompt ensembling strategy that utilizes both ChatGPT and VQA models. Firstly, we establish the criteria and guidelines for generating prompt sentences that align with our desired outcome. We have configured it to utilize the specified instructions for sentence creation. This approach necessitates the usage of attribute words to generate prompts.

Subsequently, we transmit these attribute words to ChatGPT, which leverages its large model pre-training prompt learning to automatically generate sentence prompts. Moreover, we design 
seven related questions and prompt sentence templates to cover as much information as possible about a person, aiming to gain the attributes from VQA are not included in the prompts from ChatGPT. Then the \textless question, answer\textgreater~pairs obtained by a VQA model called MCAN \cite{yu2019mcan} are converted into seven prompts corresponding to the image. This kind of prompt can be especially applied to situations when attributes cannot be obtained easily. For instance, we can ask these questions as follows: ``Is the person wearing a tie?", ``Is the person wearing a watch?", and ``What kind of shirt is the person wearing?". Next, we randomly assign questions from the question bank to each image and generate several attribute answers. We then pass these answer attributes through pre-designed declarative sentence templates and fill in the relevant words to create sub-prompt sentences. Finally, we generate ChatGPT prompts and VQA prompts for 1,501 identities in the Market1501 dataset, as well as for 1,404 identities in the DukeMTMC-reID dataset respectively.

\textbf{2) Implicit attribute prompts.}
The implicit prompt strategy in our MP-ReID method uses a learnable prompt approach that does not require intuitive attributes. Specifically, we call it ``implicit" because these learnable prompts are training dataset-specific common text descriptions, which is not \textbf{directly} corresponding to a sample. Based on CoOp \cite{zhou2022learning,zhou2022conditional} and CLIP-ReID \cite{li2022clip}, the implicit prompt mainly aims to generate concrete text descriptions through a set of learnable text tokens for fine-grained ReID tasks. That is to say, it provides some attention clues that are somewhat relevant to the tasks. For instance, let the network focus on the human body via ``the photo is a [x][x][x][x] person", not a simple/general ``the photo is a [x][x][x][x]".

\begin{table*}[]
\centering
    \label{tab:sotas}
    \renewcommand{\arraystretch}{1.1}
    \setlength{\tabcolsep}{5.25mm}{
    \begin{tabular}{ccc|cc|cc}
        \hline
            \multicolumn{3}{c|}{\textbf{Baseline}}               & \multicolumn{2}{c|}{\textbf{Market1501}}   & \multicolumn{2}{c}{\textbf{DukeMTMC-reID}} \\
            \textbf{Category}  &\textbf{Method}   & \textbf{Reference}              & \textbf{mAP}  & \textbf{R@1}            & \textbf{mAP}   & \textbf{R@1}   \\ \hline

            & SAN  & AAAI 2020         &  88.00 & 96.10    &   75.50 &   87.90     \\
              & PAT          & CVPR 2021         & 88.00 & 95.40     & 78.20   &   88.80    \\
            Image-based &  TransReID & ICCV 2021     & 88.90  &  95.20  &  82.00 &  90.70    \\
            &  MSDPA       & MM 2022          & 89.50 &  95.40 &   82.80 &  90.90       \\
            & DCAL          & CVPR 2022         & 87.50 & 94.70 &  80.10 &  89.00  \\
        \hline
            & AANet    & CVPR 2019             & 66.89 & 87.04     &  55.56 &  73.92   \\
            & AMD    & ICCV 2021             & 88.64 & 95.94    &  78.26 &  89.21    \\
            Attribute-based  & UCAD           & IJCAI 2022       &  79.50 & 92.60      &  - &  -   \\
                          
                         & UPAR     & WACV 2023       & 40.60 & 55.40   & -  & -   \\
           & CLIP-ReID     & AAAI 2023         & 89.60 & 95.50    & 82.50  & 90.00  \\
        \hline
           \textbf{Ours} & \textbf{MP-ReID} &    -      & \textbf{95.50} & \textbf{97.70}     & \textbf{88.90} & \textbf{95.70}  \\
        \hline
    \end{tabular}
    }
    \caption{Performance comparison of various state-of-the-art baselines on both datasets.}
\end{table*}

\subsection{Cross-Modal Alignment}\label{mma}
Another technical challenge is how to fuse such multi-prompts and alleviate their gaps efficiently. To address it, as shown in Figure~\ref{frameangel}, we proposed the second component of our MP-ReID --- cross-modal alignment, which eases the modality gap between textual prompt features and visual features. Furthermore, similarity learning is used to determine whether feature vectors belong to the same people or not,
\begin{equation}\label{eqn3}
sim(\mathbb{M}_{i},\mathbb{P}_{i}) = \mathbb{M}_{i} \cdot \mathbb{P}_{i} = u_M(m_{i}) \cdot u_P(p_{i})
\end{equation}
where $u_M(\cdot)$ and $u_P(\cdot)$ are linear layers projecting embedding into a cross-modal embedding space.

\textbf{1) Aligning for explicit} attribute prompts. In this module, we first perform the cross-attention operation \cite{chen2022video} with the image on both $P_{i}^{c}$ ChatGPT generation prompt and $P_{i}^{v}$ VQA generation prompt encoded by the CLIP text encoder, respectively.
Specifically, the data is sent in a sequential manner to the cross-attention module for processing. In order to integrate prompts ${f}_{i}^{p}$ and images ${f}_{i}^{m}$ more effectively, the textual prompt feature serves as query $(\boldsymbol{Q_i})$. Meanwhile, the image feature and the prompt feature perform the concatenating operation, and are subsequently utilized as key $(\boldsymbol{K_i})$ and value $(\boldsymbol{V_i})$.

Afterwards, we combine the two gained explicit prompt features, namely the ChatGPT prompt ${f}_{i}^{c}$ and the VQA prompt 
${f}_{i}^{v}$ via concatenation \cite{,zhai2022trireid}. Finally, the representation of the explicit prompts is an attentive combination of ChatGPT prompts' and VQA prompts' representations. Moreover, ${f}_{i}^{e}$ is formulated as,
\begin{equation}
{f}_{i}^{e}=MLP(Concat({f}_{i}^{c}, {f}_{i}^{v}))
\end{equation}

Then we construct a Multi-Modal Transformer model that combines prompt and image features to unify them into a cross-modal space that can be aligned \cite{luo2019bag}. After each of them receives its respective new features, the obtained features are sequentially fed into the Transformer model together with the image features, so we can get ${f}_{i}^{s}$. To further enhance the performance, we use a cross-entropy loss $\mathcal{L_{\text {cls}}}$ for the CLS token ${f}_{i}^{CLS}$, which is responsible for the classification representation of the prompts and images. $q_k$ is the value in the target distribution,
\begin{equation}
{f}_{i}^{s}=[{f}_{i}^{CLS},{f}_{i}^{e},{f}_{i}^{m}]
\end{equation}
\begin{equation}\label{eqn8}
\mathcal{L}_{cls}={\sum_{k=1}^N -q_{k}\log (MLP({f}^{s}_{i}))}
\end{equation}

We also design an image-to-prompt contrastive loss $\mathcal{L}_{m2p}$, which is calculated as follows,
\begin{equation}\label{eqn5}
\mathcal{L}_{m 2 p}(i)=-\log \frac{\exp \left(sim\left(\mathbb{M}_i, \mathbb{P}_i\right)\right)}{\sum_{a=1}^N \exp \left(sim\left(\mathbb{M}_i, \mathbb{P}_a\right)\right)}
\end{equation}

As for explicit prompt, the text-to-image contrastive loss $\mathcal{L}_{{p 2 m}}$ is formulated as,
\begin{equation}\label{eqn6}
\mathcal{L}_{{p 2 m}}(i)=-\log \frac{\exp \left(sim\left(\mathbb{M}_i, \mathbb{P}_i\right)\right)}{\sum_{a=1}^N \exp \left(sim\left(\mathbb{M}_a, \mathbb{P}_i\right)\right)}
\end{equation}

Equation (\ref{eqn5}) and Equation (\ref{eqn6}) are the similarities of two embeddings from matched pair. 

\textbf{2) Aligning for implicit} attribute prompts. As for implicit prompts, the prompts $P_{i}^{l}$ are designed as ``A photo of a $[X]_{1}[X]_{2}[X]_{3}...[X]_{T}$ person", where each $[X]_{t}$ (with $t \in {1...T}$) is a learnable text token with the same dimension as the word embedding. Here, $T$ represents the number of learnable prompt tokens. Notably, the parameters in $X$ can be trained. We can use the obtained implicit prompt features to calculate image-to-prompt cross-entropy $\mathcal{L}_{m2pce}$,
\begin{equation}\label{eqn7}
\mathcal{L}_{m 2 p c e}(i)=\sum_{k=1}^N-q_k \log \frac{\exp \left(sim\left(\mathbb{M}_i, \mathbb{P}_i \right)\right)}{\sum_{a=1}^N \exp \left(sim\left(\mathbb{M}_i, \mathbb{P}_a\right)\right)}
\end{equation}

Finally, in this module, the losses are summarized as follows,
\begin{equation}\label{eqn8}
\begin{aligned}
\mathcal{L}_{align} &=\mathcal{L}_{cls}+\mathcal{L}_{m2p}+\mathcal{L}_{p2m}+\mathcal{L}_{m2pce}
\end{aligned}
\end{equation}
\begin{table*}[]
    \renewcommand{\arraystretch}{1}
    \label{ablationsm}
    \resizebox{\textwidth}{!}{
    \begin{tabular}{cccccc|cccc|cccc}
    \hline
    \multicolumn{6}{c|}{\textbf{Strategies}}     & \multicolumn{4}{c}{\textbf{Market1501}}     & \multicolumn{4}{c}{\textbf{DukeMTMC-reID}}                    \\ 
    \hline
    \textbf{LP(Baseline)} & \textbf{AW} & \textbf{GC} & \textbf{VP} & \textbf{CP} & \textbf{CP \& VP} & \textbf{mAP} & \textbf{R@1} & \textbf{R@5} & \textbf{R@10} & \textbf{mAP} & \textbf{R@1} & \textbf{R@5} & \textbf{R@10}\\
    \hline
    
    \checkmark &             &             &             &             &             & 89.60   & 95.50    & -   & -    & 82.50    & 90.00   & -       & -   \\
    \checkmark & \checkmark  &             &             &             &             & 89.40   & 95.60    & 96.90   & 97.30    & 83.90    & 92.30   & 96.50   & 97.20   \\
    \checkmark &             & \checkmark  &             &             &             & 86.30   & 94.00    & 97.60   & 98.60    & 78.10    & 88.60   & 93.60   & 95.10   \\
    \checkmark &             &             & \checkmark  &             &             & 87.60   & 94.20    & 97.50   & 98.70    & 78.20    & 84.70   & 93.20   & 94.80   \\
    \checkmark &             &             &             & \checkmark  &             & 90.20   & 95.90    & 98.80   & 99.30    & 87.20    & 94.50   & 97.60   & 98.30   \\
    \checkmark &             &             &             &             & \checkmark  & \textbf{95.50}   & \textbf{97.70}    & \textbf{99.20}   & \textbf{99.50}    & \textbf{88.90}    & \textbf{95.70}   & \textbf{98.00}   & \textbf{98.70}   \\      
    \hline
    \end{tabular}
    }
    \caption{Ablation study of prompt strategies for MP-ReID on both datasets (Thereinto, ``LP" is learnable prompts, ``AW" is coarse and separate attribute words, ``GC" is generation caption, ``VP" is VQA prompts, ``CP" is ChatGPT prompts.}
\end{table*}

\subsection{Person Retrieval}\label{pr}
Through the above steps, we employ the Euclidean distance to calculate the distance score between query images and gallery images. Therefore, a higher score will be generated for a positive pair of person images than those of negative counterparts. In order to optimize ReID models, two loss functions are introduced: a triplet loss $\mathcal{L}_{tri}$ \cite{hermans2017defense} and an ID loss $\mathcal{L}_{id}$ \cite{zheng2017person}. The triplet loss is used to minimize the distance between images of the same person while maximizing the distance between images of different people. The ID loss, on the other hand, is used to concretely optimize for correct identity predictions by smoothing label information. By utilizing both the triplet and ID losses, the model is able to simultaneously reduce intra-class distances and increase inter-class distances, resulting in improved accuracy in re-identifying individuals,
\begin{equation}\label{eqn8}
\mathcal{L}_{id}={\sum_{k=1}^N -q_{k}\log (y_{k})}
\end{equation}
\begin{equation}\label{eqn9}
\mathcal{L}_{tri} = max(d_{p} - d_{n}+\alpha,0)
\end{equation}
where ${y}_{k}$ represents ID prediction logits of class $k$. $d_p$ and $d_n$ are feature distances of the positive and negative pair, and $\alpha$ is the margin of triplet loss.

Overall, the objective function of our method MP-ReID is denoted as follows, where $\lambda_{tri}$ is the balance factor of triplet loss and $\lambda_{id}$ is the balance factor of ID loss,
\begin{equation}\label{eqn7}
\begin{aligned}
\mathcal{L}_{reid} &=\lambda_{id} \mathcal{L}_{id}+ \lambda_{tri}\mathcal{L}_{tri}
\end{aligned}
\end{equation}

And ultimately, the loss function used in MP-ReID is as follows,
\begin{equation}\label{eqn7}
\begin{aligned}
\mathcal{L} &=\mathcal{L}_{align}+\mathcal{L}_{reid}
\end{aligned}
\end{equation}

\section{Experiments}

\subsection{Experimental Settings}\label{expd}

\subsubsection{\textbf{Dataset.}} In this paper, we evaluate the proposed MP-ReID method on two well-known ReID benchmarks: Market1501 \cite{zheng2015scalable}, DukeMTMC-reID \cite{zheng2017unlabeled}, as well as the attribute datasets associated with these two datasets, which were manually annotated \cite{lin2019improving}.

\subsubsection{\textbf{Evaluation Protocols.}} To evaluate the performance of our approach, we employed Rank@k and mean Average Precision (mAP) as the evaluation metrics, which are defined in \cite{wang2021beyond,farooq2022axm}. In particular, we applied the proposed multi-prompts learning to enhance all image features and ranked them accordingly. For all experiments on the two datasets. Higher values indicate better performance.
\subsubsection{\textbf{Implementation Details.}} 
We apply our method on a server equipped with the NVIDIA GeForce RTX 2080 Ti GPU. We use the Transformer-based models and the learning rate is $5 \times 10^{-7}$ with a linearly growing. And the warming up is set to $10$ to make the model converge faster. In our implementation, we set $S = 16$ and $K = 4$ to enable our model to learn from multiple identities and samples per identity. For feature extraction, prompt features and image features are represented as $512$-dimensional vectors. Furthermore, we set the ID loss balance factor $\lambda_{id}$ to $0.25$ as a regularization strategy, $\lambda_{tri}$ and the weight of $\mathcal{L}_{align}$ is set to $1$. Regarding the triplet loss, we set the margin parameter $\xi$ to $0.3$ to create an adequate boundary between the positive and negative samples. Moreover, we directly use the off-the-shelf ChatGPT 3.5 for the explicit prompts generation.

\subsection{Overall Performance Comparison}\label{sec:expcompare}
To demonstrate the effectiveness of our proposed method, we compared it with several state-of-the-art approaches. And we employ R@1, R@5, R@10 for convenience of representation.

Table 1 presents the experimental results, and we have the following observations: 1) Our MP-ReID approach achieves better performance on both datasets, significantly outperforming state-of-the-art baselines. It is mainly because the MP-ReID model employs the multi-prompts paradigm to significantly enhance the identification performance. This suggests the presence of highly informative cues in the image and prompt that were neglected in traditional person ReID schemes. 2) Despite recent advancements in attribute-based algorithms for person ReID, several popular methods such as AANet \cite{chen2021explainable}, AMD \cite{chen2021explainable}, UCAD \cite{sun2018beyond}, UPAR \cite{specker2023upar} and CLIP-ReID \cite{li2022clip} have demonstrated poor search results due to technological limitations that prevent full utilization of attribute information. On the other hand, image-based methods such as SAN \cite{jin2020semantics}, PAT \cite{li2021diverse}, TransReID \cite{he2021transreid}, MSDPA \cite{cheng2022more} and DCAL \cite{zhu2022dual}, while effective in some regards, do not take into account attribute information, leading to a potential loss of valuable information for person ReID. The utilization of multi-prompts in MP-ReID significantly improves the retrieval performance of person ReID. In particular, the performance of R@1 on the Market1501 dataset and the DukeMTMC-reID dataset improves significantly by at least 5.9\% and 6.1\%, respectively.

\subsection{Ablation Studies}\label{abstudy}
The overall comparative analysis shows that our proposed MP-ReID solution exhibits superior performance. To further validate the importance of multi-prompts in ReID, we took CLIP-ReID with implicit prompt as a baseline and performed some ablation studies. Firstly, MP-ReID is compared with its several variants: 1) MP-ReID with the coarse and separate attributes prompt. 2) MP-ReID with generation caption prompts from an image captioning model. 3) MP-ReID with/without any ensembling explicit prompts, i.e., ChatGPT generation prompts, as well as VQA generation prompts.

\begin{figure}[t!]
    \centering
    \label{visual1}
    \includegraphics[width=0.8\linewidth]{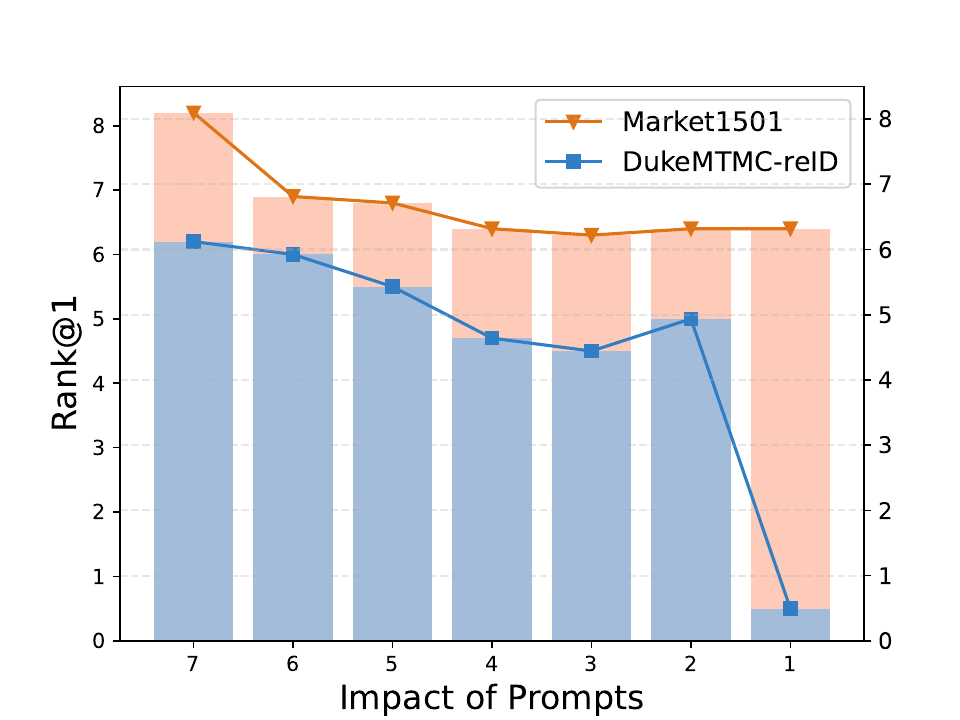}
    \caption{Ablation study on the effect of reducing various sub-prompts for MP-ReID.}
    \label{visual1}
\end{figure}

\textbf{1) Ablation study of prompt strategies for MP-ReID.} Table 2 displays the performance of different component combinations of MP-ReID. Our conclusions are threefold: a) MP-ReID using coarse and separate attribute words and generation caption shows wicked retrieval results than the prompts generated by the large model ChatGPT. b) both the explicit prompt and the implicit prompt in the table show relatively better performance. c) MP-ReID outperforms MP-ReID w/ VQA generation prompt by 7.9\% and 10.7\% in mAP on Market1501 and DukeMTMC-reID datasets. Furthermore, the scheme of MP-ReID w/ ChatGPT generation prompt proved inferior to MP-ReID by 5.3\% and 1.7\% in mAP on Market1501 and DukeMTMC-reID datasets. Furthermore, research has shown that using multiple fine prompts is more effective.

\textbf{2) Ablation study on multiple prompts.} As Figure~\ref{visual1} revealed, to further gain deeper insight into the effectiveness of multi-prompts learning in MP-ReID, we compared the effect of different numbers of multi-prompts in R@1 by showcasing on Market1501 and DukeMTMC-reID datasets. Thereinto, the graph is presented by subtracting the base value of 89\% from the obtained R@1. This presentation method is utilized to enhance the clarity of the graph. Significantly, we have the following observations: a) we use multiple sub-prompts, including 7 VQA prompts, 1 ChatGPT prompt and 1 learnable prompt. We gradually eliminate 1 - 4 VQA prompts and 1 ChatGPT prompt when only one learnable prompt remains in our experiments. The results have obviously shown that more prompts are more effective than few prompts for ReID, because few prompts cannot be learned much information. b) These findings certify the effectiveness of combining the ChatGPT generation prompts and VQA generation prompts these two explicit prompts and the implicit prompts components in our MP-ReID approach. In addition, the ablated results reveal the necessity of multiple prompts proposed in our framework, jointly resulting in its superior performance.

\begin{figure}[t!]
    \centering
    \includegraphics[width=1.0\linewidth,height=5.0cm]{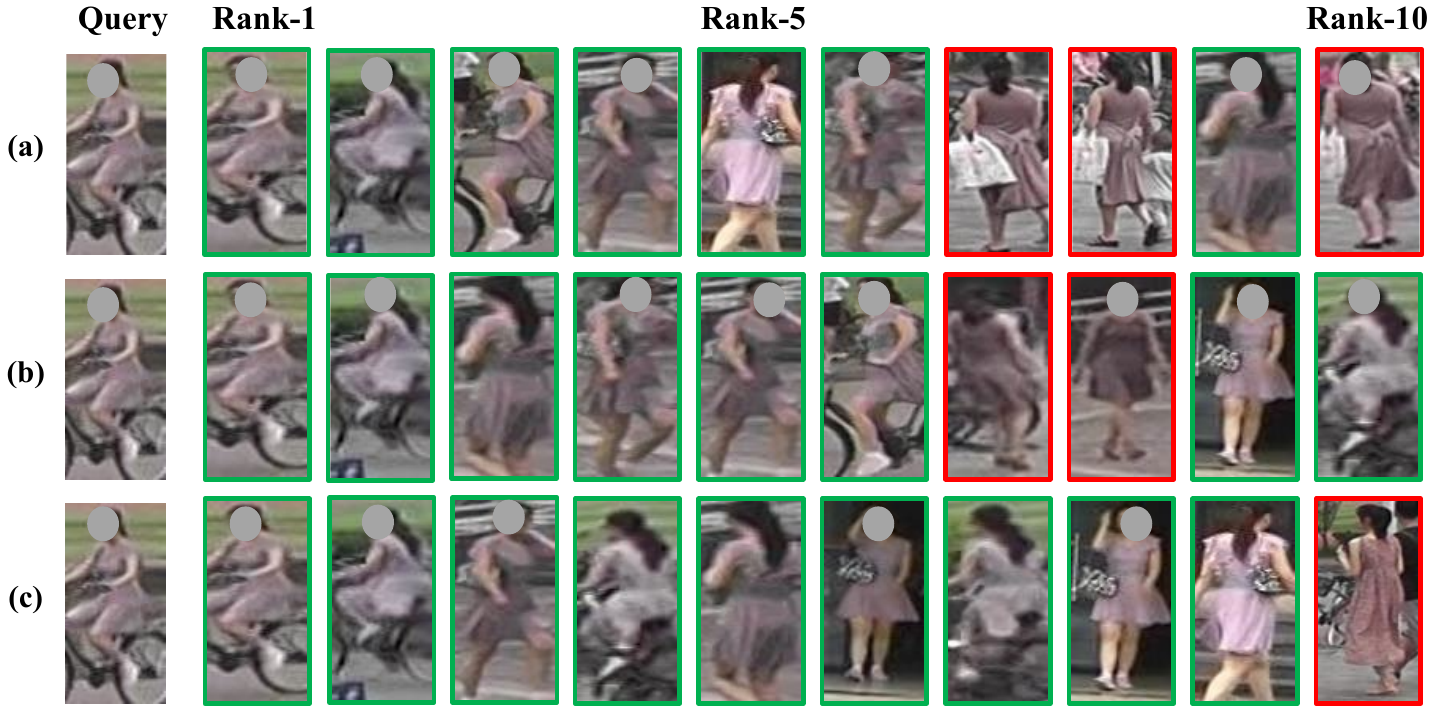}
    \caption{Visualization of three examples that illustrate the retrieval results about a) baseline (implicit learnable prompt); b) + coarse and separate attribute words; and c) our MP-ReID. Thereinto, the green box denotes the same ID as the query image, and the red box indicates a different ID from the query image.}
    \label{visual3}
\end{figure}

\subsection{Visualization} \label{visualizations}
This paper aims to combine multiple prompts features to enhance visual features. As Figure~\ref{visual3} reveals, to better comprehend our MP-ReID network, we examined and evaluated the person retrieval outcomes through visualization and analysis. Figure~\ref{visual3}a displays the effects of the baseline CLIP-ReID with implicit learnable prompts. The method adopts implicit prompts integrating coarse and separate attributes see Figure~\ref{visual3}b. And Figure~\ref{visual3}c is our MP-ReID with multiple prompts enhanced for ReID. The green box means the same ID as the query image, and the red box reveals a different ID from the query image. We can observe that MP-ReID achieves the optimal performance on Rank-10, which is mainly because of the newly introduced multi-prompts learning. Our proposed prompts learning strategies facilitate the discovery of fine-grained discriminative clues by leveraging more relevant characteristic prompts among samples.

\section{Conclusions}
This paper introduces a new concept of multi-prompts and a novel framework for attribute-based ReID, named MP-ReID. Specifically, we take the first attempt to explore multiple prompts generation learning strategies with ChatGPT and VQA models, which effectively learn discriminative representations via generated multi-prompts information. For the concrete prompts generation, we classify it into explicit prompts and implicit prompts. Among them, for generating explicit prompts, large model ChatGPT and VQA are used based on a prompt ensembling paradigm, and the implicit prompts are learnable prompts. The model is then refined using well-designed losses that consider textual prompts and visual image constraints to alleviate the modality gap. Our MP-ReID has achieved state-of-the-art performance on two well-known ReID datasets. 

\section*{Limitations}
In this paper, we propose the use of multiple prompts to enhance the person re-identification task, which has been experimentally validated as effective. However, explicit and implicit integration aspects warrant further exploration. For instance, in terms of quantity, additional prompt methods beyond the current three can be considered. Furthermore, the integration strategy can be further refined. Our current integration strategy is relatively straightforward, but we believe that employing more diverse and tighter integration methods will yield even better results. At the model level, we are particularly intrigued by multi-modal large models, but due to dataset and resource constraints, we have not yet conducted extensive experimentation. We anticipate that larger models and corpora will reveal more intriguing findings.

\section{Acknowledgments}
The authors are highly grateful to the anonymous referees for their careful reading and insightful comments. The work is supported by the National Natural Science Foundation of China (No. 61802121, No. 62302246, No. U20A20174), the Natural Science Foundation of Hunan Province, China (No. 2022JJ30159 and No. 2023JJ20013), Technology Projects of Hunan Province (No. 2015TP1004), Science and Technology Key Projects of Changsha City (No. kh2103003) and the Natural Science Foundation of Zhejiang Province, China (No. LQ23F010008) and China Scholarship Council.

\bigskip

\bibliography{aaai24}

\end{document}